\documentclass[letterpaper, 10 pt, conference]{ieeeconf}  %

\IEEEoverridecommandlockouts                              %

\usepackage{latexsym}
\usepackage{graphicx}
\usepackage{subfig}
\usepackage{amsmath}
\usepackage{amssymb}
\usepackage{blindtext}
\usepackage{todonotes}
\usepackage{wrapfig}
\usepackage{siunitx}
\graphicspath{{figures/}}
\let\labelindent\relax
\usepackage{enumitem, kantlipsum}
\usepackage{caption}
\usepackage{multirow}
\usepackage{diagbox}
\usepackage[ruled]{algorithm2e}
\usepackage{makecell}
\usepackage{tabularx}
\usepackage{rotating}
\usepackage{titlesec}
\usepackage{xcolor}
\usepackage{color, colortbl}
\usepackage{slashbox}
\usepackage{subcaption}
\usepackage{array}
\usepackage{pifont}
\usepackage{makecell}
\usepackage[hidelinks]{hyperref}

\newcommand\subfootnote[1]{%
  \begingroup
  \renewcommand\thefootnote{}\footnote{#1}%
  \addtocounter{footnote}{-1}%
  \endgroup
}
\definecolor{darkkblue}{rgb}{0.08, 0.37, 0.50}
\newcommand*\colourcheck[1]{%
  \expandafter\newcommand\csname #1check\endcsname{\textcolor{#1}{\ding{52}}}%
}
\colourcheck{blue}
\colourcheck{green}
\colourcheck{red}

\newcommand{\AlgName}{FeelAnyForce}
\title{\LARGE \bf
\AlgName{}: Estimating Contact Force Feedback
from Tactile Sensation for Vision-Based Tactile Sensors
}

\author{Amir-Hossein Shahidzadeh$^{*1}$, Gabriele Caddeo$^{*2, 3}$, Koushik Alapati$^{1}$, \\ Lorenzo Natale$^{2}$, Cornelia Ferm\"{u}ler$^{1}$, Yiannis Aloimonos$^{1}$%
\thanks{$^{1}$Perception and Robotics Group, University of Maryland College-Park}%
\thanks{$^{2}$ 
        {\tt\small b.d.researcher@ieee.org}}%
}

\definecolor{Gray}{gray}{0.9}

\begin{document}

\twocolumn[{%
\renewcommand\twocolumn[1][]{#1}%
\maketitle
\begin{center}
    \vspace{-0.1in}
    \centering
    \captionsetup{type=figure}
    \includegraphics[width=\linewidth]{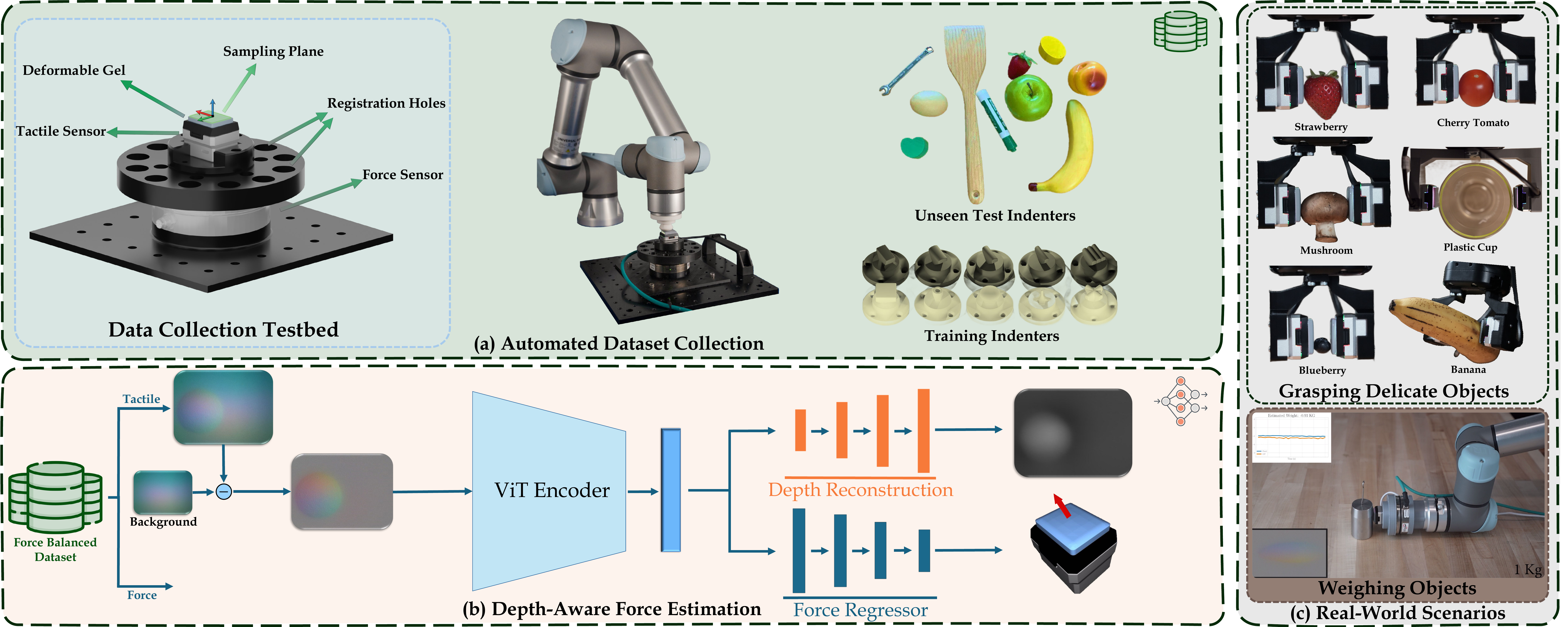}
    \begin{tikzpicture}[remember picture, overlay]
            \node at (2.15, 2.95) {\small $\mathcolor{darkkblue}{\hat{D_i}}$}; 
            \node at (-4.9, 2.6) {\small $\mathcolor{darkkblue}{T_i}$};
            \node at (-1.2, 1.10) {\small $\mathcolor{darkkblue}{x_i}$}; 
            \node at (-6.7, 0.88) {\small $\mathcolor{darkkblue}{F_i}$};
            \node at (2.15, 1.57) {\small $\mathcolor{darkkblue}{\hat{F_i}}$};
        \end{tikzpicture}
        \vspace{-3mm}
    \captionof{figure}{We present \AlgName{}, a method for estimating contact forces with sensor generalization capabilities on vision-based tactile sensors. (a) We collect a dataset of tactile-depth-force data by using a robotic arm to press various indenters onto a tactile sensor mounted on a force sensor. We then test its performance on a set of YCB and real world objects. (b) To isolate the contact data $T_i$, we subtract the sensor specific background image. The network is trained to minimize the Force regression error along with depth reconstruction loss. Note that the ground-truth depth $D_i$ is computed using photometric stereo in Gelsight mini. (c) We showcase real-world experiments conducted with our force estimator.}
    \label{fig:scheme}
    \vspace{-0.1in}
\end{center}
}]

\subfootnote{
\hspace{-3mm}\textsuperscript{$1$} University of Maryland, College-Park, MD, USA \\
\textsuperscript{$2$}  Istituto Italiano di Tecnologia, Via San Quirico, 19 D, Genova, Italy. \\
\textsuperscript{$3$} DIBRIS, Universit\`a di Genova, Via All'Opera Pia, 13, Genoa, Italy. \\
\textsuperscript{*} Equal contributions. \\
The support of NSF under awards OISE 2020624 and BCS 2318255 is greatly acknowledged.
}
\thispagestyle{empty}
\pagestyle{empty}

\begin{abstract}
  In this paper, we tackle the problem of estimating 3D contact forces using vision-based tactile sensors. In particular, our goal is to estimate contact forces over a large range (up to 15 N) on any objects while generalizing across different vision-based tactile sensors. Thus, we collected a dataset of over 200K indentations using a robotic arm that pressed various indenters onto a GelSight Mini sensor mounted on a force sensor and then used the data to train a multi-head transformer for force regression. Strong generalization is achieved via accurate data collection and multi-objective optimization that leverages depth contact images.
  Despite being trained only on primitive shapes and textures, the regressor achieves a mean absolute error of 4\%  on a dataset of unseen real-world objects. We further evaluate our approach's generalization capability to other GelSight mini and DIGIT sensors, and propose a reproducible calibration procedure for adapting the pre-trained model to other vision-based sensors.
Furthermore, the method was evaluated on real-world tasks, including weighing objects and controlling the deformation of delicate objects, which relies on accurate force feedback. Supplementary material and extended discussions are available at \href{http://prg.cs.umd.edu/FeelAnyForce}{{\color{blue}http://prg.cs.umd.edu/FeelAnyForce}}
\end{abstract}

\section{Introduction}

Through the sense of touch, humans interact with precise force to manipulate objects dexterously, adapting to object features. Having accurate force control while interacting with different objects is important during manipulation or when dealing with fragile or deformable objects. 
Using their hands, humans can simultaneously perceive other fine-grained quantities of the external world, such as shapes, textures, and hardness. No single robotics platform can replicate this multi-modality perception~\cite{LUO201754,reviewtactileinfo}, but research in tactile sensing is advancing. Various tactile sensor technologies have been developed to mimic human perception, yet no single technology stands out as superior, as each has its limitations. 
Nowadays, vision-based tactile sensors \cite{gelsight,digit,tactip} have become the state-of-the-art in the field, reaching high performance in tasks such as shape reconstruction \cite{suresh2022shapemap,shahidzadeh2023actexplore}, texture recognition \cite{texture} and local geometry estimation \cite{caddeocollision,caddeo2023sim2real,suresh2022midastouch}, thanks to high-resolution output. However, mapping such high-resolution output to a continuous value of force, along with sensor specific non-linear mechanical properties of the gel and non-uniform light distribution, makes force estimation a non-trivial problem. While adding markers can partially solve this problem, it leads to lower performance when perceiving other object characteristics. Moreover, no prior work with tactile sensors demonstrates sufficient precision and generalization across real-world objects of varying shapes, textures and over a wide range of forces, while achieving consistent performance across different types of sensors. \\
The goal of this paper is to build a 3D force estimator for vision-based tactile sensors able to generalize to \emph{unseen objects} and to \emph{different vision-based tactile sensors} with high accuracy on a large range of forces (up to 15 N). To solve this problem, we use the GelSight Mini sensor, which outputs RGB tactile and depth images. In particular, we collect a large dataset (over 200k indentations) of calibrated real-world tuples of tactile images, depth images, and forces on 10 3D-printed primitive shapes (indenters). The diversity of indenters is chosen to help the network focus on the meaningful features without overfitting on specific shapes. The dataset was collected by varying the 6D relative pose between the sensor and the indenters with different indentation levels, measured through careful design and calibration of an automated data collection setup. We then use this dataset to train a multi-head architecture consisting of a Vision Transformer~\cite{dosovitskiy2021image} (ViT) backbone pre-trained on DINOv2~\cite{oquab2024dinov2} followed by a regressor head that estimates 3D forces and a depth-reconstruction head. This approach, combined with the robust features from pre-trained weights and the accurate dataset, allows the ViT to focus on the force-related characteristics of images, leading to better generalization on unseen objects and sensors. We validate the robustness of the architecture on a set of unseen everyday-life objects and YCB model objects \cite{7254318}. Furthermore, we propose a fast fine-tuning calibration procedure through a 3D-printable setup and a set of off-the-shelf weights to obtain better force estimation performance on a specific combination of sensor and gel. We also test the same procedure on the DIGIT sensor obtaining promising results. %
To demonstrate the applications of such an accurate tactile-force estimator, we validate the performances on two downstream tasks: \emph{weighing objects by pushing} and \emph{controlled deformation of delicate objects}, such as plastic cups or fruits.
We summarize our contributions as follows:
\begin{itemize}
    \item A depth-aware force estimation architecture capable of adapting to unseen vision-based sensors in both static and dynamic conditions.
    \item A large labeled dataset on 10 indenters including tactile images, depth images, 3D forces vectors.
    \item A low-cost, user-friendly calibration procedure to obtain comparable performances on different vision-based tactile sensors.
\end{itemize}

\section{Related works}
\subsection{Tactile Force Estimation}
Tactile Force estimation has attracted significant attention in recent research among various types of sensors to extends the tactile sensors capabilities.
Capacitive sensors were employed in \cite{icub} where the force estimation performance was only assessed on a single indenter and within limited range of forces.
In \cite{sundaralingam2019robust}, a network was trained to predict 3D forces using the BioTac sensor while performing grasp and place tasks. However,  the BioTac sensor's limited force range and frequent calibration requirements can hinder its practical application. In \cite{softbubble}, the authors used a FEM-based numerical method to estimate force fields on a soft bubble sensor. However, this approach faces challenges in maintaining accuracy under high deformation conditions. Recently, vision-based tactile sensors have become more prominent \cite{doi:10.1080/01691864.2019.1632222}, producing high-resolution images by capturing gel deformations from contact \cite{reviewcamera}. \cite{designmotivation} achieved robust performance on marker-based tactile sensors with FEM-based methods on a single spherical indenter and limited forces. Likewise, the GelSlim sensor \cite{gelslim} and the biomimetic sensor in \cite{vitactip} were tested on limited ranges of forces and objects. On the other hand, \cite{biotactip} validated with 10 indenters achieving comparable absolute error to ours on a limited force range. Nevertheless
, markers can interfere with tasks like shape reconstruction, prompting research into marker-free methods with better generalization capabilities across different sensors. CNN-based methods were applied on GelSight \cite{gelsight}, showing difficulties in handling various indenters' shapes not present in training. \cite{lf3} modifies the GelSight sensor design to make it more suitable for force estimation. They tested their approach on a significant range of forces, but the post-calibration force error on different sensors of a kind remains noticeable. In \cite{tact9d}, the authors demonstrated their new sensor estimating forces-torque on unseen 3D-printed objects. They validated their approach on unseen objects but their model's ability to generalize across different sensors remains unexplored. Our work estimates 3D force vectors from GelSight Mini, training on diverse shapes and textures. We achieve state-of-the-art force estimation accuracy on real-world and YCB objects, demonstrating reliable performance across a wide force range and various sensors.

\subsection{Tactile Feedback in Control and Manipulation}

Tactile force feedback implemented in a closed-loop approach is crucial for grasping and manipulation tasks, as performance is limited without it \cite{sundaresan2023learning,ganguly2022gradtac}. While force sensors remain common in state-of-the-art methods \cite{shaikewitz2023inmouth}, vision-based tactile sensors have recently proven useful for control and manipulation by inferring other valuable information beyond force.
Some works used vision-based tactile feedback for control without directly relying on force. For instance, \cite{tian2019manipulation} presented a learning-based tactile MPC framework based on tactile images, while others used it to build manipulation controllers \cite{hogan2020tactile}, \cite{sunil2022visuotactile}.
These sensors were also applied to grasping tasks for evaluating success \cite{calandra2017feeling}, stability \cite{si2022grasp}, or slip detection \cite{dong2018maintaining}, \cite{james2020slip}. In \cite{kolamuri2021improving}, model-based tactile feature extraction methods were used to develop grasping controllers.
Our work aims to enhance the multimodal capabilities of vision-based tactile sensors by incorporating reliable force estimation into their skill set.

\section{Method}
In the following, we explain our dataset collection procedure (Sec. \ref{subsec:dataset}, Fig. \ref{fig:scheme} a). Then, we present our model for estimating 3D forces (Sec. \ref{subsec:model}, Fig. \ref{fig:scheme} b). Finally, we employ this regression model for two challenging real-world tasks: weighing objects by pushing and interacting with delicate objects (Sec. \ref{subsec:tasks}, Fig \ref{fig:scheme} c).

\subsection{Dataset Collection}\label{subsec:dataset}

\begin{figure}
\vspace{2mm}\includegraphics[width=0.48\textwidth]{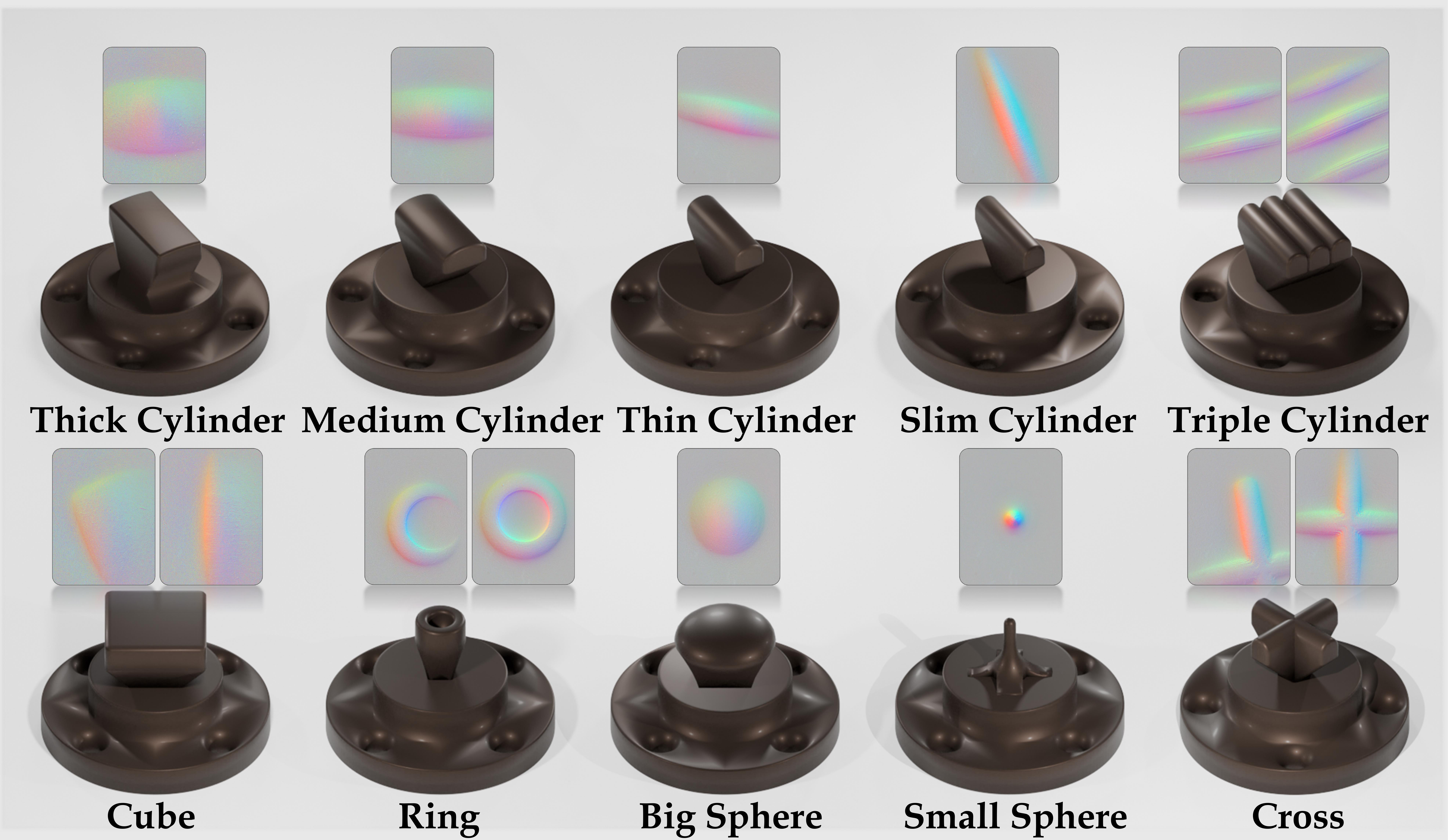}
    \caption{The indenters used for the training procedure.}
    \label{fig:indenter}
    
\end{figure}

We require our dataset to have ground-truth net force vectors $(\in \mathbb{R}^3)$ for various indentations on tactile sensors. To this end, we design a testbed (Fig. \ref{fig:scheme}) to affix the tactile sensor on a force sensor and use a robotic arm to drive the indenters on the tactile sensor from different poses. To align the robot's end-effector pose ($\mathcal{R}$) with the sensor's coordinate system ($\mathcal{S}$), a set of key points is added to the testbed allowing to find the transformation between them using a 3-point registration algorithm \cite{7750718}. To replicate indentations of real-world objects, we use a set of primitive indenters inspired by objects humans manipulate daily (Fig. \ref{fig:indenter}). We want them to be sufficiently descriptive of the world around us. The indenters not only comprise different shapes and curvatures, but also separate contact patches (triple cylinder), non-convex contact (ring and cross), and flat surfaces (cube). 
To generate indentation trajectories driven by the arm, we first move the end-effector to the sensor's coordinate system origin $O_S$ using the transformation discussed earlier. Subsequently, we define a sampling plane $\mathcal{P}$ where $z_\mathcal{S}=0$ (see Fig. \ref{fig:scheme}). Then we sample a $(x, y)$ pair from $\mathcal{P}$, and we experimentally find a range of safe orientations $(\theta, \zeta, \eta)$ based on an indenter's shape to avoid breaking the sensor. Specifically, for every indenter, we would have a set of indentation poses: 
\begin{align*}
    \{(x,y,0,\alpha, \beta, \gamma) | x &\in [-X,+X], y \in [-Y, +Y], \\ \alpha &\in [-\theta, \theta], \beta \in [-\zeta, \zeta], \gamma \in [-\eta, \eta]\} 
\end{align*}

After sampling the indentation poses ($\in \mathbb{SO}^3$), we perform indentations by gradually moving the indenter in small steps along the z-axis of the end-effector's coordinate system $\mathcal{R}$ determined by sampled end-effector Euler angle rotations $(\alpha, \beta, \gamma)$. By moving along the end-effector's z-axis, we achieve angled indentations that result in various force angles in the dataset. To synchronize the tactile sensor and force measurements and obtain the highest resolution from the force sensor (0.04 N), it is necessary to allow the sensors to stabilize after each indentation step. This is made possible using a robotic automated data collection approach. It would not be feasible in manual data collection because of the challenge of keeping the end-effector steady, which would result in noisy force reading. We collecte over \textbf{200K} samples of tactile-depth-force vectors using 10 3D-printed primitive indenters for training as demonstrated in Fig. \ref{fig:indenter} . For evaluation, we use a subset of YCB and some real world objects (Fig. \ref{fig:scheme} a) to represent the error on real-world unseen objects. Due to sensor characteristics,the dataset has a bias toward higher $F^z$ (more than 15N) compared to $F^x, F^y$ (range of 4N) due to the sensor's flat surface, which cannot tolerate higher shear forces. After collecting data with all the indenters, there would be a non-uniform distribution for every force range. To eliminate this bias, we balance the dataset  for every indenter in a way that we get a uniform histogram for most force ranges covered by the indenters (more discussions and distribution in Supp. Materials).

\subsection{Network Design} 
\label{subsec:model}
\vspace{-0.05cm}
We use a ViT-based~\cite{dosovitskiy2021image} network pretrained on DINOv2 \cite{oquab2024dinov2} as encoder. The zero-shot robustness of the features across domains helps the model focus on force-related image characteristics, reducing the risk of overfitting to the training domain. Moreover, force depends on the area and gel deformation. For this reason leveraging the information given by the associated depth images can better represent the tactile-force relationship. Therefore, we design a multi-head architecture with a regressor and a decoder. The first outputs the 3D force vector, while the latter outputs the estimated depth image to condition the encoder. Using the depth as labels and not as input is necessary to make the network compatible with sensors that don't output the depth, e.g., the DIGIT sensor. The network  takes as input RGB images $T_i \in \mathbb{R}^{240 \times 320 \times 3}$ from the GelSight mini sensor (padded to $\mathbb{R}^{320 \times 320 \times 3}$, rescaled to $\mathbb{R}^{224 \times 224 \times 3}$) and outputs a feature vector $x_i \in \mathbb{R}^{K \times 1}$. We subtract the no-contact tactile background image from other tactile images to enhance the network's robustness to sensor production inconsistencies and to help the model focus on contact patches. For the depth images, we find the $min_{val}$ and $max_{val}$ of the train set, and we normalize the images between [$min_{val} - \epsilon$, $max_{val} + \epsilon$] to avoid saturation of out-of-distribution samples.
The latent vector $x_i$ encoded by the ViT is then fed into the regressor and the decoder. Specifically, the regressor is an MLP-head made of four bottleneck layers (a linear layer, followed by Layer Normalization \cite{ba2016layer} and a GELU \cite{hendrycks2016gaussian} activation function) and a final linear layer which outputs the estimated force $\hat{F_i} \in \mathbb{R}^{3 \times 1}$. The decoder is composed of 4 deconvolutional layers followed by a Leaky ReLU activation function and outputs the estimated depth $\hat{D_i} \in \mathbb{R}^{224 \times 224 \times 1}$. To assess the effectiveness of our design choices, we conducted various ablation studies discussed in Sec. \ref{sec:experiments}.

\textbf{Model training}. We sample a batch of training data ($T_i$, $D_i$, $F_i$) with size $N$. We define $\phi(\cdot)$ to be our ViT feature extractor, $\beta(\cdot)$ our bottleneck layers, $\rho(\cdot)$ our linear regressor, and $\psi(\cdot)$ our depth-reconstruction decoder. We compute the predicted force of the regressor $\hat{F_i}$ and the depth image of the decoder $\hat{D_i}$ as follows:
\begin{align}
    \hat{F_i} &=  \rho(\beta(\phi(T_i))) \\
    \hat{D_i} &= \psi(\phi(T_i)) 
\end{align}
and we train the network to minimize loss $\mathcal{L}$ as follows:
\begin{align}
    \mathcal{L}_F(F_i,\hat{F_i}) &= \frac{1}{N} \left|\hat{F_i}-F_i \right|_1 \\
    \mathcal{L}_D(D_i,\hat{D_i}) &= \frac{1}{N}\left\Vert \hat{D_i} - D_i \right\Vert_2 \\
    \mathcal{L}(\mathcal{L}_F, \mathcal{L}_D) &= \alpha  \mathcal{L}_F +  \beta  \mathcal{L}_D 
\end{align}

\subsection{Downstream tasks}
\label{subsec:tasks}
\textbf{Object weighing by pushing}. We apply our estimator to perform object weighing by pushing. As shown in Fig. \ref{fig:Pushing}, we mount the GelSight mini tactile sensor on top of the  ATI-Axia80 force sensor to get the ground truth, and we affix this setup on the robot arm to drive the pushing. We assume that the pressure distribution of the sensor is uniform, and so is the mass density distribution of the object, meaning the center of mass $c$ is in the center. Moreover, we assume the push force $f_p$ to be parallel to the ground and intersect the center of mass $c$ to avoid any rotation component in the computation. We assume the object translates in a straight line. Given an object $O$ with mass \textit{m} moving with linear velocity \textit{v}
\begin{equation}
    f_{p} = m\dot{v} + f_d,
\end{equation}
where $f_d = \mu mg$, with $\mu$ the dynamic friction coefficient between the object and the ground, and $g$ is the gravity acceleration. Moving the robot with a constant velocity allows us to weigh the object as $mg = \frac{f_p}{\mu}$.

\textbf{Controlled deformation of delicate objects}. The second task is the controlled deformation of deformable objects. We perform quantitative and qualitative experiments. The goal of the quantitative experiment is to control the deformation of a plastic cup with a gripper by applying a force estimated by two GelSight Mini sensors mounted on the gripper. This experiment involves estimating the required forces to achieve desired deformations. We use the setup described in the weighing experiments to determine the ground truths. One tactile sensor is mounted on the robot arm on top of the force sensor, while the other is affixed to the ground. The interaction between the object, placed on top of the second tactile sensor, and the arm is conducted by incrementally moving the robotic arm downwards under quasi-static conditions. We simulate a parallel-jaw grasp by aligning tactile sensors pushing against each other. The force measured by the force sensor is recorded when the object expands to the desired percentage. This force value is applied by both tactile sensors, assuming the same contact area. The deformation percentage is measured by fitting an ellipse to the plastic cup's rim using classical image processing methods in OpenCV \cite{opencv_library}, treating the cup as a circle when not in contact and as an ellipse when push force is applied. This approach makes the deformation computation agnostic to the relative distance to the camera as we compute deformation relative to the un-deformed state. The chosen contact area is the rim of the plastic cup, which can be easily treated and assimilated to a circle in an unloaded state. This procedure is repeated 10 times, with slight orientations between trials to account for minor variations in the contact area. We average the recorded forces to determine the thresholds. Additionally, the goal of qualitative results is to stably grasp delicate fruits while preventing physical damage on , as shown in Fig \ref{fig:scheme}c.

\section{Experimental Results}
\label{sec:experiments}
We validate our model by testing its accuracy performance on static interactions against objects and on two downstream tasks using force-tactile feedback, a dynamic object weighing by pushing, and controlled deformation of deformable objects. The model used for these experiments is trained for $100$ epochs with Adam optimizer and with a learning rate of \num{5e-5} for the backbone and \num{1e-5} for the heads, and we refer to it as DINO RD.

\begin{table*}
\vspace{0.7em}
\centering
\caption{L1 normalized error on unseen objects of the different architectures. \label{tab:accuracy}}
\resizebox{2\columnwidth}{!}{
\begin{tabular}{ l | c | c | c | c | c | c | c | c | c | c| c }

Method & Banana & Brown egg & Strawberry & Marker & Wooden spatula & Ellipsoid & Nectarine & Green apple & Wrench & Heart & Mean \\
\hline

ResNet RD & 4.0\% & 5.4\% & 5.9\%  & 5.2\% & 7.2\% & 5.0\% & 4.5\% &  \textbf{4.7\%} & 3.3\%& 7.8\%& 5.3\% \\
\rowcolor{Gray} DINO$_{F}$ RD & 6.3\% & 6.9\%& 13.2\% & 11.3\% & 6.7\% & 6.8\% & \textbf{3.4\%} & 5.8\% & 15\% & 9.2\% & 8.5\% \\
DINO R & 4.9\% & 6.3\% & 5.1\% & \textbf{3.6\%} & 5.4\% & 5.3\% & 5.2\% & 6.1\% & \textbf{2.3\%} & 7.6\% & 5.2\% \\
\rowcolor{Gray} DINO RD & \textbf{3.5\%} & \textbf{4.5\%} & \textbf{3.0\%} & 4.3\% & \textbf{4.9\%} & \textbf{3.4\%} & 4.3\% & 5.0\% & 3.1\%  & \textbf{5.8\% } & \textbf{4.2\% }\\
\end{tabular}
}
\end{table*}

\vspace{-0.7em}
\subsection{Static force estimation}
\label{sec:result_static}
We perform static force estimation accuracy tests by comparing the performance of the proposed approach with other architectures to validate our design choices on unseen objects. The unseen objects dataset comprises everyday-life objects like YCB objects that could not be attached to the robot. We mounted a Robotiq Hand-e gripper on the UR5 robotic arm to grasp the object and touch the sensor with random relative poses, collecting the data once the contacts were stable. The tested forces cover all the ranges for the different axes. We validate the choice of the encoder and its fine-tuning by replacing it with a pre-trained ResNet RegressorDecoder (ResNet RD) and by training the architecture with the ViT weights frozen (DINO$_F$ R), respectively. Moreover, we show the role of the multi-head architecture by comparing it with a network without the decoder (DINO R). Table \ref{tab:accuracy} highlights the accuracy on unseen objects for the models considered. The proposed architecture reaches an average error of 4.3\% over the unseen objects, outperforming the other methods by at least 20\% of relative erorr. Moreover, the performance stays always between 3\%-5.8\%, showing a high robustness across the different shapes and textures. In particular, the performance on objects with complex textures such as the strawberry and the wooden spatula or  difficult shapes such as the upper part of the heart highlights the robustness of the method when compared with the other approaches. The DINO$_F$ RD obtains significantly higher normalized error with respect to the other methods, showing how important is the finetuning for the tactile images domain. ResNet RD and DINO R achieve good performance, assessing the quality of the train dataset collection. However, they present a notable drop of performance when compared to the proposed method. Furthermore, we can see how their accuracies vary on a higher range (between 3.3\%-7.8\% and 2.3\%-7.6\% respectively), showing less generalization capabilities. 
To further validate our approach, we test the performance of the estimator on a balanced dataset of 700 images in a range up to 15 N on different GelSight Mini sensors and gels. We use the sensor and gel used during data collection (Sensor 1 and Gel 1) and two more sensors and gels (Sensor 2-3 and Gel 2-3). For this test, the images are collected using the indenters to precisely test the performance with the same relative indenter-sensor poses.
We consider all combinations of three gels and three sensors, as shown in Table~\ref{tab:generalization}. While DINO$_F$ R consistently presents higher errors under every condition, DINO R demonstrates similar performance on the training sensor when varying the gels compared to DINO RD. This result suggests that the network can achieve outstanding results on the training data and sensor, even when the gels are changed. In contrast, the performance on other sensors highlights the importance of the depth-reconstruction head. Similarly, ResNet RD achieves the best performance on the training sensor and gel with the seen objects dataset, but struggles with other combinations, showing a high dependency on the gel's mechanical properties and the sensor's illumination. This behavior further emphasizes the role of the ViT backbone. On the other hand, the proposed method maintains remarkable stability across all combinations, consistently keeping the error between 4.2\% and 6.3\%, despite differences between sensors and gels. This robustness across different sensors enables the use of the same network for downstream tasks described in Sec. \ref{exp:tasks}, where multiple sensors are required.

\begin{table}[ht]
    \vspace{-2mm}
    \centering
    
            \caption{Normalized error of the networks for the train and test sensors varying the gels.}
            \label{tab:generalization}
            \resizebox{0.48\textwidth}{!}{
        \begin{tabular}{c|c|c|c|c|c|c|c|c|c|c}
                & \multicolumn{3}{c|}{\textbf{Sensor 1 (train)}} & \multicolumn{3}{c|}{\textbf{Sensor 2}} & \multicolumn{3}{c|}{\textbf{Sensor 3}} & \\
                \cline{2-10}
                
                 \backslashbox{\textbf{Method}}{\textbf{Gel}} & 1  & 2 & 3 & 1  & 2 & 3 & 1  & 2 & 3 & \textbf{Mean}\\
        \Xhline{2.5\arrayrulewidth}
        ResNet RD & \textbf{3.9\%} & 5.2\% & 6.4\% & 7.7\% & 5.6\%& \textbf{5.1\%} & 7.2\% & \textbf{5.8\%} & 7.3\% & 6.0\% \\
        \rowcolor{Gray} DINO${_F}$ R & 7.4\% & 10.5\% & 7.4\% & 11.8\% & 8.3\% & 8.5\%& 7.8\% & 9.8\%& 8.0\% & 8.8\% \\
        DINO R & 4.9\% & \textbf{4.3\%} & \textbf{4.5\%} & 8.0\% & 5.8\%& 5.5\%& 6.6\% & 6.3\%& 6.4\% & 5.8\%\\
        \rowcolor{Gray} DINO RD & 4.2\% & 4.8\% & 5.2\% & \textbf{6.3\%} & \textbf{4.8\%} & \textbf{5.1\%} & \textbf{6.1\%} & 6.3\%& \textbf{6.2\%} & \textbf{5.4\%} \\
            \end{tabular}}

\end{table}
\vspace{-2mm}
\subsection{Calibration}
\label{exp:calibration}

    To compensate for sensor-specific manufacturing variations, we propose a reproducible calibration procedure using a 3D printable setup and a set of off-the-shelf weights, with more details available in the supplementary material. The setup consists of a bottom mount to fix the sensor and a rotatable upper mount with a fixed indenter. By placing the weights in known positions on the upper mount, users can collect images with known forces, registered via our force sensor. In Tab. \ref{tab:digit} we test the performance of this approach collecting 100 images with sensor 2 and sensor 3 using respectively gel 1 and gel 2 along with a DIGIT sensor, and we test on the same dataset collected for those combinations used for Tab. \ref{tab:generalization}. The 100 images are collected using the Big Sphere indenter shown in Fig. \ref{fig:indenter}, intentionally not included in this test dataset. For the Gelsight sensors, the results show that our method reaches performance comparable to those obtained by training combination of sensor and gel, while the same cannot be said for the other methods. Moreover, after calibration, our method maintains nearly the same accuracy on the uncalibrated sensors, with only a negligible average error increment of 4\% in the normalized error on those sensors, compared to the 25\%, 14\%, and 12\% increases observed for ResNet RD, DINO$_F$, and DINO R, respectively (e.g., a 10\% increment of 5\% is 5.5\%). Regarding the DIGIT sensor, due to the relevant differences in light distribution and mechanical properties, we need to finetune the entire regressor head. The high error rate of the ResNet-based architecture suggests that it struggles to generalize across different sensor architectures. For the ViT, the results confirm that conditioning the backbone with a depth-reconstruction head is crucial for achieving good generalization. Specifically, DINO R exhibited triple the error compared to our approach, and double the error compared to the frozen backbone. This suggests that the depth-reconstruction head conditions the encoder to learn \textbf{force-related} features rather than \textbf{sensor specific} features. We further demonstrate this characteristic by comparing the mean of the attention layers of the three different encoders in the attached video.
\begin{table}[ht]
    \vspace{-1mm}
    \centering

            \caption{Normalized error of the networks for unseen GelSight Mini sensors and DIGIT sensor after finetuning.}
            \label{tab:digit}
        \begin{tabular}{c|c|c|c|c}
                & ResNet RD & DINO${_F}$ R & DINO R & DINO RD\\
                
        \Xhline{2.5\arrayrulewidth}
         Sensor 2 - Gel 1 & 4.7\% & 7.2\% & 5.4\% & \textbf{4.2\%} \\
         \rowcolor{Gray} Sensor 3 - Gel 2 & 4.9\% & 7.9\% & 6.0\% & \textbf{4.3\%} \\
         DIGIT & 53.4\% & 16.0\% & 29.6\% & \textbf{9.2\%} \\
            \end{tabular}
    \vspace{-3mm}

\end{table}

\vspace{-1mm}
\subsection{Downstream tasks}
\label{exp:tasks}

\begin{figure}[t]
        \vspace{5mm}
        \centering
        \vspace{-0.25cm}
        \includegraphics[width=0.46\textwidth]{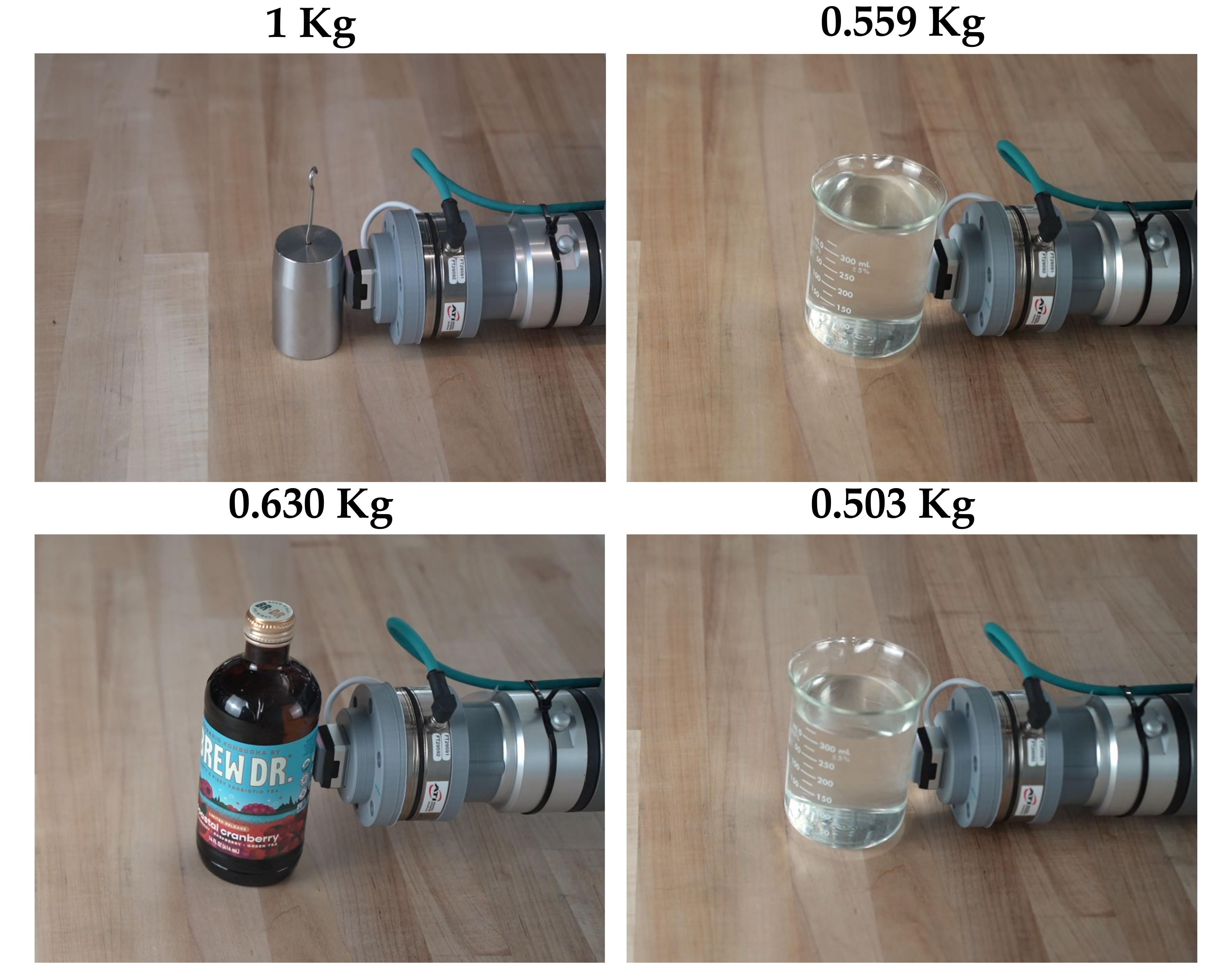}
        \caption{We show sampled frames from trajectories of weighing by pushing. We consider objects that differ in weight, shape, and material.}
        \label{fig:Pushing}
\end{figure}

\textbf{Object weighing by pushing}. 
For the weighing experiments, we utilized the sensor employed during the training procedure. We consider three different objects, shown in Fig. \ref{fig:Pushing}: a 1Kg calibration weight made of metal, a caliber cylinder full of water (0.559 Kg) or almost full (0.503 Kg), and a bottle of glass of 0.63 Kg. These objects differ in contact shape, material, texture, and weight. Five pushes of approximately 15 seconds each per object are conducted at a controlled low velocity to minimize errors from the wooden table surface's imperfections and reduce the object's rotational component, which is not considered in the estimation. The $z$ component of the estimated force is averaged for all experiments for both the GelSight and the force sensor, and the force error is computed by subtracting the two averages. The dynamic friction value is determined by using the force sensor's measurements as the ground truth. Table \ref{tab:weight} reports the forces estimated by the tactile sensor and force sensor, along with the corresponding error measured and subsequent weight estimated. The network shows good performance on weight estimation ranging from 6\% to 10\%. This result further validates the need for a high-precision estimator, demonstrating how differences of just 0.06N can affect the downstream tasks.
\begin{table}[ht]
    \vspace{-2mm}
    \centering
                \caption{Results of weighing objects by pushing experiments for the four objects considered. }
            \label{tab:weight}
    \resizebox{0.48\textwidth}{!}{
    
        \begin{tabular}{c|c|c|c|c}
            
                {weight[Kg]} & \textbf{Weight(1)} & \textbf{Cyl.(0.503)}  & \textbf{Cyl.(0.559)} & \textbf{Bottle(0.630)}\\ 
                \hline

        \Xhline{2.5\arrayrulewidth}
        Force GelSight (Our) [N] & 2.05 &0.87  & 0.98 & 1.02\\
        \rowcolor{Gray} Force ATI (GT) [N] & 2.24 & 0.96 & 1.06 & 1.08\\
        Force error [N] & 0.2 & 0.09 & 0.09 & 0.06\\
        \rowcolor{Gray} Weight error [Kg] & 0.1 & 0.04  & 0.05 & 0.04\\
            \end{tabular}}

\end{table}

It is important to point out some non-idealities that contribute to the estimation error. The friction value is fitted to match the force estimation from the force sensor, but the force sensor itself is also subject to errors. The force sensor tends to drift slightly towards higher values over time, leading to increased error as the push time increases since the tactile sensors consistently predicted lower forces than the force sensor. This behaviour is more evident when the force sensor is in motion, as in this task. Unlike the data collection, where the force sensor was tared before every contact, in this task is not possible to tare the force sensor for all the pushing period, leading to a higher drift of forces. However, the accuracy is consistent regardless the instrument error, the texture or the contact surface shape, which varies from flat surface to curves with different radii of curvature, not seen during the training procedure. Further details are shown in the video.

\begin{figure}[t]
        \centering
        \vspace{2mm}\includegraphics[width=0.47\textwidth]{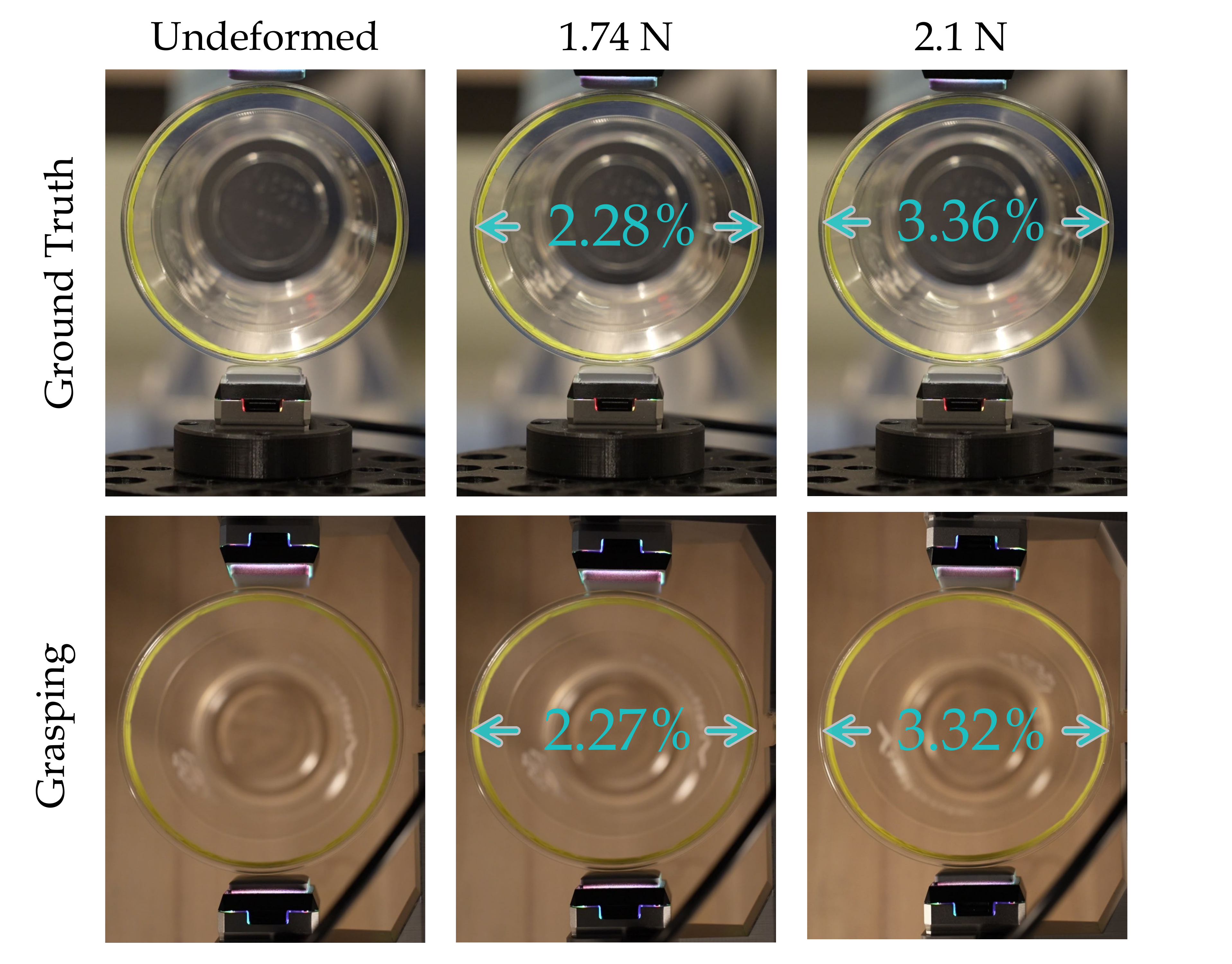}
        \caption{Grasping experiments with a plastic cup. We close the gripper to get $f_2=1.74$ N and $f_3= 2.1$ N from the force sensor and our method in the first and second row, respectively.}
        \label{fig:deformation}
        \vspace{-4mm}
\end{figure}

\textbf{Grasping of delicate objects.}
Fig. \ref{fig:deformation} highlights the quantitative results of the interaction experiments with the plastic cup. We gradually close the gripper on the rim of the cup until either of the sensors reaches the target force in the $z$ component, then measure the percentage of deformation. This experiment is repeated 10 times, and the results are averaged. The ground truth values show deformations of 2.28\% and 3.36\%, while the tactile estimator measures deformations of 2.27\% and 3.32\% for the two target forces, with a deformation error of less than 0.05\%. It is important to note that the discrete steps of the gripper limit the experiment's precision, as also highlighted in \cite{piga}. Therefore, the desired force could not always be reached, and the closest step to the desired value is chosen. Additionally, the rim is a shape different from those seen in the training set (slimmer), further validating the generalization capabilities of our approach. 
Regarding the qualitative experiments, we empirically determine two force thresholds: gentle grasp (minimum force to hold an object) and firm grasp (maximum force before damage). We tested these thresholds on banana, tomato, mushroom, and strawberry, finding specific values for each. We gradually closed the gripper, guided by tactile sensor feedback (see Fig. \ref{fig:scheme}), to reach these thresholds. We recorded success rates and provide extensive details in the accompanying video.

\section{Limitations and
Conclusions}
\label{sec:conclusion}
\textbf{Limitations}. While the presented results demonstrate state-of-the-art accuracy in static and dynamic force estimation for vision-based tactile sensors, there are still areas for improvement. As outlined in Sec. \ref{subsec:dataset}, there is a significant variance in sensing range across axes. This characteristic emanates from the flat and soft gel which cannot tolerate higher shear forces. Moreover, even if the approach can generalize on different GelSight Mini sensors and achieve optimal performance thorugh the calibration procedure, it cannot reach the same performance on other vision-based tactile sensors with limited data, even if the results are promising. We aim to tackle the problem in future works.

\textbf{Conclusions}. In \AlgName, we introduced the first force estimator for vision-based tactile sensors capable of generalizing on different sensors achieving high accuracy regardless of shape and texture on a large range of forces. We meticulously collected a real-world dataset of associated forces, RGB images, and depth images sampled through the interaction with specific indenters. Leveraging RGB-D tactile images from the GelSight Mini sensor, our transformer-based force estimator demonstrated exceptional performance in accurately inferring 3D force vectors. We demonstrated robust performance of the model in both static and dynamic conditions through comprehensive experiments and ablation studies. Furthermore, our investigations into the model's adaptability to different GelSight Mini sensors demonstrate that the community can readily utilize the model, with promising results on DIGIT sensors. We release model, and data, hoping our work can assist the community in advancing research in multimodal tactile sensing.  

\bibliographystyle{IEEEtran}
\bibliography{root}

\begin{thebibliography}{10}
\providecommand{\url}[1]{#1}
\csname url@samestyle\endcsname
\providecommand{\newblock}{\relax}
\providecommand{\bibinfo}[2]{#2}
\providecommand{\BIBentrySTDinterwordspacing}{\spaceskip=0pt\relax}
\providecommand{\BIBentryALTinterwordstretchfactor}{4}
\providecommand{\BIBentryALTinterwordspacing}{\spaceskip=\fontdimen2\font plus
\BIBentryALTinterwordstretchfactor\fontdimen3\font minus \fontdimen4\font\relax}
\providecommand{\BIBforeignlanguage}[2]{{%
\expandafter\ifx\csname l@#1\endcsname\relax
\typeout{** WARNING: IEEEtran.bst: No hyphenation pattern has been}%
\typeout{** loaded for the language `#1'. Using the pattern for}%
\typeout{** the default language instead.}%
\else
\language=\csname l@#1\endcsname
\fi
#2}}
\providecommand{\BIBdecl}{\relax}
\BIBdecl

\bibitem{LUO201754}
\BIBentryALTinterwordspacing
S.~Luo, J.~Bimbo, R.~Dahiya, and H.~Liu, ``Robotic tactile perception of object properties: A review,'' \emph{Mechatronics}, vol.~48, pp. 54--67, 2017. [Online]. Available: \url{https://www.sciencedirect.com/science/article/pii/S0957415817301575}
\BIBentrySTDinterwordspacing

\bibitem{reviewtactileinfo}
Q.~li, O.~Kroemer, Z.~Su, F.~Veiga, M.~Kaboli, and H.~Ritter, ``A review of tactile information: Perception and action through touch,'' \emph{IEEE Transactions on Robotics}, vol.~PP, pp. 1--16, 07 2020.

\bibitem{gelsight}
\BIBentryALTinterwordspacing
W.~Yuan, S.~Dong, and E.~H. Adelson, ``Gelsight: High-resolution robot tactile sensors for estimating geometry and force,'' \emph{Sensors}, vol.~17, no.~12, 2017. [Online]. Available: \url{https://www.mdpi.com/1424-8220/17/12/2762}
\BIBentrySTDinterwordspacing

\bibitem{digit}
\BIBentryALTinterwordspacing
M.~Lambeta, P.-W. Chou, S.~Tian, B.~Yang, B.~Maloon, V.~R. Most, D.~Stroud, R.~Santos, A.~Byagowi, G.~Kammerer, D.~Jayaraman, and R.~Calandra, ``Digit: A novel design for a low-cost compact high-resolution tactile sensor with application to in-hand manipulation,'' \emph{IEEE Robotics and Automation Letters}, vol.~5, no.~3, p. 3838–3845, Jul. 2020. [Online]. Available: \url{http://dx.doi.org/10.1109/LRA.2020.2977257}
\BIBentrySTDinterwordspacing

\bibitem{tactip}
\BIBentryALTinterwordspacing
B.~Ward-Cherrier, N.~Pestell, L.~Cramphorn, B.~Winstone, M.~E. Giannaccini, J.~Rossiter, and N.~F. Lepora, ``The tactip family: Soft optical tactile sensors with 3d-printed biomimetic morphologies,'' \emph{Soft Robotics}, vol.~5, no.~2, pp. 216--227, 2018, pMID: 29297773. [Online]. Available: \url{https://doi.org/10.1089/soro.2017.0052}
\BIBentrySTDinterwordspacing

\bibitem{suresh2022shapemap}
S.~Suresh, Z.~Si, J.~G. Mangelson, W.~Yuan, and M.~Kaess, ``Shapemap 3-d: Efficient shape mapping through dense touch and vision,'' 2022.

\bibitem{shahidzadeh2023actexplore}
A.-H. Shahidzadeh, S.~J. Yoo, P.~Mantripragada, C.~D. Singh, C.~Fermüller, and Y.~Aloimonos, ``Actexplore: Active tactile exploration on unknown objects,'' 2023.

\bibitem{texture}
\BIBentryALTinterwordspacing
A.~Böhm, T.~Schneider, B.~Belousov, A.~Kshirsagar, L.~Lin, K.~Doerschner, K.~Drewing, C.~A. Rothkopf, and J.~Peters, ``What matters for active texture recognition with vision-based tactile sensors,'' 2024. [Online]. Available: \url{https://arxiv.org/abs/2403.13701}
\BIBentrySTDinterwordspacing

\bibitem{caddeocollision}
G.~M. Caddeo, N.~A. Piga, F.~Bottarel, and L.~Natale, ``Collision-aware in-hand 6d object pose estimation using multiple vision-based tactile sensors,'' in \emph{2023 IEEE International Conference on Robotics and Automation (ICRA)}, 2023, pp. 719--725.

\bibitem{caddeo2023sim2real}
G.~M. Caddeo, A.~Maracani, P.~D. Alfano, N.~A. Piga, L.~Rosasco, and L.~Natale, ``Sim2real bilevel adaptation for object surface classification using vision-based tactile sensors,'' 2023.

\bibitem{suresh2022midastouch}
S.~Suresh, Z.~Si, S.~Anderson, M.~Kaess, and M.~Mukadam, ``Midastouch: Monte-carlo inference over distributions across sliding touch,'' 2022.

\bibitem{dosovitskiy2021image}
A.~Dosovitskiy, L.~Beyer, A.~Kolesnikov, D.~Weissenborn, X.~Zhai, T.~Unterthiner, M.~Dehghani, M.~Minderer, G.~Heigold, S.~Gelly, J.~Uszkoreit, and N.~Houlsby, ``An image is worth 16x16 words: Transformers for image recognition at scale,'' 2021.

\bibitem{oquab2024dinov2}
M.~Oquab, T.~Darcet, T.~Moutakanni, H.~Vo, M.~Szafraniec, V.~Khalidov, P.~Fernandez, D.~Haziza, F.~Massa, A.~El-Nouby, M.~Assran, N.~Ballas, W.~Galuba, R.~Howes, P.-Y. Huang, S.-W. Li, I.~Misra, M.~Rabbat, V.~Sharma, G.~Synnaeve, H.~Xu, H.~Jegou, J.~Mairal, P.~Labatut, A.~Joulin, and P.~Bojanowski, ``Dinov2: Learning robust visual features without supervision,'' 2024.

\bibitem{7254318}
B.~Calli, A.~Walsman, A.~Singh, S.~Srinivasa, P.~Abbeel, and A.~M. Dollar, ``Benchmarking in manipulation research: Using the yale-cmu-berkeley object and model set,'' \emph{IEEE Robotics \& Automation Magazine}, vol.~22, no.~3, pp. 36--52, 2015.

\bibitem{icub}
N.~Jamali, M.~Maggiali, F.~Giovannini, G.~Metta, and L.~Natale, ``A new design of a fingertip for the icub hand,'' in \emph{2015 IEEE/RSJ International Conference on Intelligent Robots and Systems (IROS)}, 2015, pp. 2705--2710.

\bibitem{sundaralingam2019robust}
B.~Sundaralingam, A.~Lambert, A.~Handa, B.~Boots, T.~Hermans, S.~Birchfield, N.~Ratliff, and D.~Fox, ``Robust learning of tactile force estimation through robot interaction,'' 2019.

\bibitem{softbubble}
J.-C. Peng, S.~Yao, and K.~Hauser, ``3d force and contact estimation for a soft-bubble visuotactile sensor using fem,'' 2023.

\bibitem{doi:10.1080/01691864.2019.1632222}
\BIBentryALTinterwordspacing
A.~Yamaguchi and C.~G. Atkeson, ``Recent progress in tactile sensing and sensors for robotic manipulation: can we turn tactile sensing into vision?1,'' \emph{Advanced Robotics}, vol.~33, no.~14, pp. 661--673, 2019. [Online]. Available: \url{https://doi.org/10.1080/01691864.2019.1632222}
\BIBentrySTDinterwordspacing

\bibitem{reviewcamera}
\BIBentryALTinterwordspacing
K.~Shimonomura, ``Tactile image sensors employing camera: A review,'' \emph{Sensors}, vol.~19, no.~18, 2019. [Online]. Available: \url{https://www.mdpi.com/1424-8220/19/18/3933}
\BIBentrySTDinterwordspacing

\bibitem{designmotivation}
\BIBentryALTinterwordspacing
C.~Sferrazza and R.~D’Andrea, ``Design, motivation and evaluation of a full-resolution optical tactile sensor,'' \emph{Sensors}, vol.~19, no.~4, 2019. [Online]. Available: \url{https://www.mdpi.com/1424-8220/19/4/928}
\BIBentrySTDinterwordspacing

\bibitem{gelslim}
D.~Ma, E.~Donlon, S.~Dong, and A.~Rodriguez, ``Dense tactile force estimation using gelslim and inverse fem,'' in \emph{2019 International Conference on Robotics and Automation (ICRA)}, 2019, pp. 5418--5424.

\bibitem{vitactip}
W.~Fan, H.~Li, W.~Si, S.~Luo, N.~Lepora, and D.~Zhang, ``Vitactip: Design and verification of a novel biomimetic physical vision-tactile fusion sensor,'' 2024.

\bibitem{biotactip}
H.~Li, S.~Nam, Z.~Lu, C.~Yang, E.~Psomopoulou, and N.~F. Lepora, ``Biotactip: A soft biomimetic optical tactile sensor for efficient 3d contact localization and 3d force estimation,'' \emph{IEEE Robotics and Automation Letters}, vol.~9, no.~6, pp. 5314--5321, 2024.

\bibitem{lf3}
W.~Li, M.~Wang, J.~Li, Y.~Su, D.~Jha, X.~Qian, K.~Althoefer, and H.~Liu, ``L$^3$ f-touch: A wireless gelsight with decoupled tactile and three-axis force sensing,'' \emph{IEEE Robotics and Automation Letters}, vol.~PP, p. 2023, 07 2023.

\bibitem{tact9d}
C.~Lin, H.~Zhang, J.~Xu, L.~Wu, and H.~Xu, ``9dtact: A compact vision-based tactile sensor for accurate 3d shape reconstruction and generalizable 6d force estimation,'' 2023.

\bibitem{sundaresan2023learning}
P.~Sundaresan, J.~Wu, and D.~Sadigh, ``Learning sequential acquisition policies for robot-assisted feeding,'' 2023.

\bibitem{ganguly2022gradtac}
K.~Ganguly, P.~Mantripragada, C.~M. Parameshwara, C.~Ferm{\"u}ller, N.~J. Sanket, and Y.~Aloimonos, ``Gradtac: Spatio-temporal gradient based tactile sensing,'' \emph{Frontiers in Robotics and AI}, vol.~9, p. 898075, 2022.

\bibitem{shaikewitz2023inmouth}
L.~Shaikewitz, Y.~Wu, S.~Belkhale, J.~Grannen, P.~Sundaresan, and D.~Sadigh, ``In-mouth robotic bite transfer with visual and haptic sensing,'' 2023.

\bibitem{tian2019manipulation}
S.~Tian, F.~Ebert, D.~Jayaraman, M.~Mudigonda, C.~Finn, R.~Calandra, and S.~Levine, ``Manipulation by feel: Touch-based control with deep predictive models,'' 2019.

\bibitem{hogan2020tactile}
F.~R. Hogan, J.~Ballester, S.~Dong, and A.~Rodriguez, ``Tactile dexterity: Manipulation primitives with tactile feedback,'' 2020.

\bibitem{sunil2022visuotactile}
N.~Sunil, S.~Wang, Y.~She, E.~Adelson, and A.~Rodriguez, ``Visuotactile affordances for cloth manipulation with local control,'' 2022.

\bibitem{calandra2017feeling}
R.~Calandra, A.~Owens, M.~Upadhyaya, W.~Yuan, J.~Lin, E.~H. Adelson, and S.~Levine, ``The feeling of success: Does touch sensing help predict grasp outcomes?'' 2017.

\bibitem{si2022grasp}
Z.~Si, Z.~Zhu, A.~Agarwal, S.~Anderson, and W.~Yuan, ``Grasp stability prediction with sim-to-real transfer from tactile sensing,'' 2022.

\bibitem{dong2018maintaining}
S.~Dong, D.~Ma, E.~Donlon, and A.~Rodriguez, ``Maintaining grasps within slipping bound by monitoring incipient slip,'' 2018.

\bibitem{james2020slip}
J.~W. James and N.~F. Lepora, ``Slip detection for grasp stabilisation with a multi-fingered tactile robot hand,'' 2020.

\bibitem{kolamuri2021improving}
R.~Kolamuri, Z.~Si, Y.~Zhang, A.~Agarwal, and W.~Yuan, ``Improving grasp stability with rotation measurement from tactile sensing,'' 2021.

\bibitem{7750718}
J.~A. Marvel and K.~Van~Wyk, ``Simplified framework for robot coordinate registration for manufacturing applications,'' in \emph{2016 IEEE International Symposium on Assembly and Manufacturing (ISAM)}, 2016, pp. 56--63.

\bibitem{ba2016layer}
J.~L. Ba, J.~R. Kiros, and G.~E. Hinton, ``Layer normalization,'' \emph{arXiv preprint arXiv:1607.06450}, 2016.

\bibitem{hendrycks2016gaussian}
D.~Hendrycks and K.~Gimpel, ``Gaussian error linear units (gelus),'' \emph{arXiv preprint arXiv:1606.08415}, 2016.

\bibitem{opencv_library}
G.~Bradski, ``{The OpenCV Library},'' \emph{Dr. Dobb's Journal of Software Tools}, 2000.

\bibitem{piga}
N.~A. Piga and L.~Natale, ``Adaptive tactile force control in a parallel gripper with low positioning resolution,'' \emph{IEEE Robotics and Automation Letters}, vol.~8, no.~9, pp. 5544--5551, 2023.

\end{thebibliography}

\end{document}